\documentclass[a4paper]{amsart}

\usepackage{amsfonts,amssymb,amsmath}
\usepackage{latexsym}
\usepackage{color}
\usepackage{epsfig} 
\usepackage{geometry}
\usepackage{mathrsfs}
\usepackage{todonotes}
\usepackage[foot]{amsaddr}
\usepackage[normalem]{ulem}

\numberwithin{equation}{section}

\newcommand{\defi}{\textit}

\title{A simulated annealing approach to optimal storing in a multi-level warehouse}

\author{Alexander Eckrot}
\address{Institut f\"ur Theoretische Physik, Universit\"at Regensburg, Universit\"atsstrasse 31, 93053 Regensburg, Germany}
\email{alexander.eckrot@physik.uni-regensburg.de}

\author{Carina Geldhauser}
\address{Weierstrass Institute for Applied Analysis and Stochastics, Mohrenstrasse 39, 10117 Berlin, Germany}
\email{geldhauser@wias-berlin.de}

\author{Jan Jurczyk}
\address{Institut f\"ur Theoretische Physik, Universit\"at Regensburg, Universit\"atsstrasse 31, 93053 Regensburg, Germany}
\email{jan.jurczyk@physik.uni-regensburg.de}

\begin{document}

\begin{abstract}
  We propose a simulated annealing algorithm specifically tailored to optimise total retrieval times in a multi-level warehouse under complex pre-batched picking constraints. Experiments on real data   from a picker-to-parts order picking process in the warehouse of a European manufacturer show that optimal storage assignments do not necessarily display features presumed in heuristics, such as clustering of positively correlated items or ordering of items by picking frequency. 
   In an experiment run on more than 4000 batched orders with 1 to 150 items per batch, the storage assignment suggested by the algorithm produces  a 21\% reduction in the total retrieval time with respect to a frequency-based storage assignment. 
\end{abstract}

%A Theory and Methodology Paper classified into the  heading Production, Manufacturing and Logistics of EJOR.

\maketitle

\section{Introduction}
Warehouses play a key role in modern supply chains \cite{frazelle2002supply} and are a significant cost factor to a company: according to the European Logistics Association/AT Kearney report \cite{european2004differentiation}, the capital and operation costs of warehouses represent about 25\% of the surveyed companies' logistics costs in 2003, while figures for the USA \cite{establish} indicate that warehousing contributed to the total logistics costs with a share of 22\%.

\defi{Order picking}, generally defined as the process of retrieving products from storage in response to a specific customer request, is the most labour-intensive operation in warehouses with manual systems, and a very capital-intensive operation in warehouses with automated systems \cite{goetschalckx1989classification, tompkins2003facilities}. Estimates of the percentage of order picking costs on the total warehouse costs range as high as 55\% in Drury \cite{drury1988towards} and  Bartholdi et al\cite{bartholdi2011warehouse} to 65\% in Coyle et al. \cite{coyle1996management}. For these reasons, warehousing professionals consider order picking as the highest priority area for productivity improvements.

%https://books.google.de/books?hl=en&lr=&id=-xBIq6Qm2SQC&oi=fnd&pg=PA1&dq=Tompkins,+J.A.,+White,+J.A.,+Bozer,+Y.A.,+Frazelle,+E.H.,+Tanchoco,+J.M.A.,+Facilities+Planning&ots=sB6DndFeX2&sig=24Czdpy4kn4ZPWGfWooJK7L1WIs#v=onepage&q=Tompkins%2C%20J.A.%2C%20White%2C%20J.A.%2C%20Bozer%2C%20Y.A.%2C%20Frazelle%2C%20E.H.%2C%20Tanchoco%2C%20J.M.A.%2C%20Facilities%20Planning&f=false

Over the last twenty years, many papers have studied order picking processes and optimal strategies or heuristics to optimise subprocesses such as warehouse layout, storage assignment, order batching, order release method and picker routing. However, there is little published research on how to combine these subprocesses optimally: we mention here  \cite{de1999efficient}, which compares the performance of S-shaped and Largest Gap routing heuristics for batches of 3 and 4 items, and \cite{Hsieh2011618}, which focuses on heuristics for order batching to improve the overall performance of order picking systems.

The contribution of this work is to propose a combined optimisation of the storage assignments and a routing heuristic for correlated \defi{batched orders} of large size (up to 102 unique parts in our tests). This is achieved via a hand-tailored algorithm based on a simulated annealing combinatorial search, which incorporates  variable parameters depending on the routing heuristics. The algorithm is designed for multi-level warehouses, the routing heuristics is based on a warehouse design with wide aisles, however, it can be adapted to narrow aisles. 

\section{Warehouse order picking}

In human-operated warehouses, the most common system for order picking is the \defi{picker-to-parts} class, where the (human) order picker walks or drives along the aisles to pick items \cite{de2004assess, van1999models}. Despite their ubiquity, pickers-to-part order-picking systems have not received comparable attention from researchers, perhaps because of their variety and complexity \cite{de2007design}.

An optimal strategy for a \defi{picker-to-parts} picking process will minimize the time needed for picking all orders in a given time frame. The precise quantity to minimize is indeed \defi{total retrieval time}, which is defined as  the sum of pick and travel time and time due to delays. By using s-shaped routing or other routing rules which make the aisles unidirectional, the delay part (which is usually due to congestion) can be a priori removed from the model. The main components to be minimized are therefore (1) the time needed for travelling the warehouse to collect the items  and (2) the time needed to perform one pick. If we assume that  the microdesign of the storage racks was already optimised and appropriate tools are used by the picker, the latter is basically a constant, its size depending on the level in the warehouse where the item is located. 

Minimizing the time needed for travelling the warehouse can be achieved by various ansatzes, the most common ones involve solving  the \defi{storage location assignment problem} (SLAP), the  \defi{routing problem} or allocating optimal \defi{batched orders} for one tour of the picker.

 %%%%%%
 
\subsection{Storage assignment policies}

Early scientific contributions to the  storage location assignment problem in warehouses include a taxonomy of possible storage location assignment policies, where the classification between dedicated storage, randomized storage and class-based storage was introduced, see \cite{frazele1989correlated, frazelle1989stock, hausman1976optimal} and the references therein.

In our problem, we deal with a warehouse where the individual pieces used in the production of certain products are stored. Therefore, the orders arrive already pre-batched, according to the necessity of the assembly line. The order batches are quite large  (on average 30 items per batch in the data we used for testing our algorithm). We  therefore seek a solution to the combined SLAP for a warehouse with  \defi{dedicated storage}-policy and the routing problem for such pre-batched orders.

%We first point out some of the relevant literature in this field. An extensive literature review of design and control of warehouse order picking was made in deKoster et al \cite{de2007design}.

The \defi{dedicated storage} policy assigns a fixed location to each product, meaning that this location is reserved even for products that are out of stock. While this may result in a low space utilisation, which may imply higher maintenance costs and possibly longer routing times, dedicated storage locations are an advantage in warehouses for production units, since every stored item is needed regularly.

Moreover, dedicated storage locations help to increase the orientation of the order pickers, leading to an increased routing velocity and fewer wrong picks.  Dedicated storage policies allow for logical grouping of storage items, which are often advantageous in retail warehouses \cite{de2001logistics} and in general for items of very different weight, which can be stored in decreasing weight along the standard picking route, implying a good stacking sequence.

However, as pointed out in the survey \cite{de2007design}, analytical models for optimising dedicated %and class-based
storage assignment in manual-pick order-picking systems are still lacking. Existing studies % in picker-to-parts order-picking systems
mainly focus on random storage assignments.

The \defi{random storage policy} (see for example \cite{petersen1997evaluation}) reduces the total space required. This does not automatically improve routing times, as travel distance might increase, see %Choe and Sharp 
\cite{tutorial}. It needs a computer-controlled environment to be efficient, as a lack of automation or technical equipment assisting the picking process can lead to slow travelling times due to disorientation and higher percentages of picking errors.

%Achtung: In Zitation von deKoster fehlen Koautoren, sonst stimmt das file:  J.P. van den Berg, G.P. Sharp, Forward-reserve allocation
%in a warehouse, European Journal of Operations Research 111 (1) (1998) 98√ê113

It was observed \cite{van1998forward, van1999models} that a separation of the pick stock (forward area) from the bulk stock (reserve area) can lead to a significant improvement of picking times: in the forward area, a dedicated storage policy is applied, while the bulk area can follow a random storage policy. In this way, the advantages of dedicated storage still hold and disadvantages are reduced. Indeed, this policy is already adapted in many warehouses attached to production units, such as the one in our reference case.

%The  \defi{full-turnover storage policy} distributes products over the storage
%area according to their turnover. The products with the highest sales rates are located at the easiest accessible locations and slow moving products are stored in areas with high traveling times.

%\todo[inline]{missing class-based storage. Understand difference to grouping of products}.

%The size of the forward area is restricted: the smaller %the area, the lower the average travel times of the %order pickers will be. It is important to decide %how much of each SKU is placed in the forward %area and where in the area it has to be located} % SKU = stock keeping unit

At the interface between research and industry, several papers, such as \cite{battista2011storage, brynzer, chan2011improving, dekker2004improving, kutzelnigg}, describe algorithms that solve the SLAP with real-life constraints. In particular the paper \cite{chan2011improving} also deals with a multi-level warehouse situation, pointing out that the storage assignment systems need to reflect the structure of the orders.

\subsection{Routing policies}

Storage assignment has an impact on the performance of the routing method \cite{petersen1999}.
%However, this effect seems to be largely neglected in the literature. %Instead, many authors focus on random storage %assignment to discuss about the performance of %routing methods.
The problem of optimal routing for order picking classifies as a Steiner Traveling Salesman Problem, which is in general not solvable in polynomial time. However, for a special warehouse aisle configuration, Ratliff and Rosenthal \cite{ratliff1983order} showed that there does exist an algorithm that can solve the problem in running time linear in the number of aisles and the number of pick locations. This algorithm was extended to other situations in \cite{de1998routing,roodbergen2001routingmiddle, roodbergen2001routingmultiple}.
% by De Koster and Van der Poort \cite{de1998routing} and Roodbergen and De Koster \cite{roodbergen2001routingmiddle, roodbergen2001routingmultiple}.

However, algorithms are not yet available for every specific layout, and there remains the unsolved problem of aisle congestion by pickers following different routes, and the fact that pickers may deviate from routes that they deem illogical \cite{gademann2005order}. Because of this, the problem of routing order pickers is mainly solved by using a heuristic, such as the s-shape method. In \cite{henn2013245},  a  routing strategy for a warehouse with  U-shaped layout has been introduced and proven to be more efficient under certain conditions.

Most heuristic methods for routing order pickers in single-block warehouses assume that the aisles of the warehouse are narrow enough to allow the order picker to retrieve products from both sides of the aisle without changing position \cite{hall1993distance, petersen1997evaluation, roodbergen2001layout}. A polynomial-time algorithm for routing order pickers in wide aisles was proposed in \cite{goetschalckx1988efficient}.

One of the strengths of the algorithm we propose here is that it gives efficient heuristics  both for narrow and wide aisles: the routing strategy, the right-hand rule and the aisle length are incorporated as variable parameters and can easily be changed, see formula \eqref{eq:time_routing}.

\subsection{Simulated annealing}

The \defi{simulated annealing} (SA) method is an optimisation algorithm introduced by Kirkpatrick in 1983 \cite{kirkpatrick1984optimization} and separately by {\v{C}}ern{\`y} in 1985 \cite{vcerny1985thermodynamical}.  The idea behind this method is to emulate the congelation of a crystal with many atoms. This process starts at very high temperatures at which all atoms are free to move. By slowly decreasing the temperature atoms start to "feel" the presence of others and arrange themselves into a structure, more precisely: the interaction between atoms forces them to settle into a crystal shape, where the overall interaction energy is minimized. The slow cooling process allows the system to explore many configurations and enabling it to find a near optimal or optimal configuration.

Transferring this method to optimisation problems, one has to define an \defi{energy} or \defi{cost function} $f$. In our case this cost function is the overall picking time in a warehouse. Changing the positions of items (also called \defi{configuration}) leads to an increase or decrease in the cost function. At each iteration of this changing procedure (called \defi{update}), the simulated annealing algorithm checks whether the new configuration is better or worse than the one before and applies the Metropolis criterion \cite{metropolis1953equation}, which states the following: An improvement, in this case a change of configuration which leads to a decrease of the cost function, will be accepted, while a worse configuration will only be adopted with a \defi{Boltzmann probability}
\begin{equation}
	\mathbb{P}(\Delta f,T)=e^{-\frac{\Delta f}{T}}
\end{equation}
where $\Delta f$ is the difference of the values of the cost-function for the two configurations and $T$ a control parameter similar to a temperature.

When performing an optimisation, at first one chooses a large control parameter $T$ and performs a finite number of updates to the configuration. Then the parameter $T$ is slightly decreased and again a finite number of updates are realised.
Repeating the last step 
transforms the update-method from a random walk at large $T$ to a local search update at small $T$. The last configuration obtained is the one with the least energy.

Simulated annealing algorithms are nowadays widely used in economics, for example in packaging problems \cite{gomes2006solving}, the production scheduling problem \cite{loukil2007multi}, the corridor allocation problem \cite{ahonen2014simulated} \cite{amaral2012corridor}  and solving the SLAP \cite{atmaca2013defining, muppani2008efficient}.

A simulated annealing heuristic to minimize total retrieval time involving order batching and sequencing was introduced in \cite{hong2012batch}. Their algorithm uses a geometric cooling schedule \cite{cohn1999simulated} %(Cohn \& Fielding, 1999),
also adapted in \cite{matusiak2014fast}. The algorithm proposed in \cite{matusiak2014fast} uses a single flexible heuristic based on random moves in a structured manner, in comparison to multiple deterministic neighbourhood search heuristics, as often found in the literature, and is comparably very fast.

Recent results \cite{henn2013metaheuristics} show that solutions obtained by \defi{simulated annealing}, \defi{Iterated Local Search} or the \defi{Attribute-Based Hill Climber} \cite{Whittley2004}
%presents two metaheuristics to minimize the total tardiness for a given set of customer orders, one based on the other inspired by  %, a heuristic based on asimple tabu search principle.
may allow order picking systems to operate more efficiently compared to those obtained with standard constructive heuristics such as the \defi{Earliest Due Date rule}.

\section{Problem description and modelling}

%The algorithm we propose is based on the following model:

%We consider a production site of a medium-size European company, offering highly customized products. The individual parts for the product are pre-fabriced by subcontractors and shipped to the company's warehouse. Incoming orders to the warehouse are collected in batches according to the necessities of the assembly line (the complete batch of  \textcolor{blue}{individual items} is needed in one step of the assemblying process).

We consider  a multi-level warehouse of a production site. Incoming orders consist of large batches of individual items, which  are picked manually into a bin using a picker-to-parts order picking system and delivered to a couple of output locations. The first two levels of the warehouse are easily accessible, while all levels above can only be accessed by a lifting device, resulting in a higher picking time per item.

The items ordered in a batch are correlated to each other, as each batch corresponds to the parts used in one step of the assembly line of the production site. 

The difficulty of this SLAP is due to the high number of possible optional features among which the customer can choose when ordering  one of the (very few) main products: The number of items which are needed for all options in a main product line is relatively low compared to the number of items corresponding to one or more options.
The correlations between individual items to be picked are therefore quite complex and cannot be treated by a pure frequency-based algorithm as proposed in \cite{battista2011storage}.

Another level of complexity is added by different storage container classes used in the warehouse: As they are of different size, they cannot be randomly exchanged. Moreover, the number of different items stored per aisle depends on the composition of container classes in this aisle. Note that standard simulated annealing algorithms are unable to deal with combinatorial constraints as posed by storage containers of different size.

\section{Solution approach}

We  build a simulated annealing algorithm which finds the configuration of warehouse items in a multi-level, multi-container class warehouse minimizing the total retrieval time of batched orders under a given routing in a reference time frame.  
For this, we need to specify (1) how we calculate the retrieval time  and (2) which update routine we implement in our SA algorithm to gradually improve retrieval times.

\subsection{Construction of retrieval time}
As we consider pre-batched orders, meaning several items collected into the same bin during one route, we have to calculate the retrieval times per ordered bin. The total retrieval time $t$ in the reference time frame is simply the sum over the individual retrieval times $ t_{bin, i}$ for each bin (labelled by $i$) ordered in the reference time frame:

\begin{equation}\label{eq:totaltime}
  t =\sum_{i=1}^{N_{\text{bin}}} t_{\text{bin}, i}
\end{equation}

As mentioned above, our algorithm achieves this by both optimising the storage assignment and using a heuristic to minimize routing times.
The retrieval time per bin $t_{\text{bin}}$ is therefore split into the pick time
%\footnote{Please note that times which are calculated in each step by the algorithm, such as the routing time, are denoted by a small $t$, while ``parameter times'', such as the time the picker needs to change from one aisle to the next one (called $\tau_{\text{aisle}}$ below) are denoted by $\tau$. Times denoted by $\tau$ are not calculated but inserted into the calculation as parameters. They need to be measured in the individual warehouse to which the algorithm is applied.} 
$t_p$ and the routing time $t_r$. These are calculated independently from one another and the retrieval time per bin then reads
\begin{equation}
  t_{\text{bin}} = t_p + t_r.
\end{equation}

\subsubsection{Routing heuristic}
The \defi{routing time} is the time that a picker needs to physically move to all locations in the warehouse where the goods for one bin are stored. The goal is to reduce the overall routing time for all bins.

It is easy to see that for a warehouse with only parallel aisles, the s-shaped routing heuristic can be applied both for wide aisles, enforcing a right-hand pick rule, and for narrow aisles (no right-hand rule). However, while unidirectional narrow aisles have to be traveled completely, the routing heuristic in wide aisles would be to pick with the right hand until the last item in the batch for the right part of the aisle is reached, and then to turn around and pick from the ``left'' part of the aisle, exiting at the same point where the aisle was entered.

We present a routing heuristic optimised for wide aisles. This formula can readily be adapted to an s-shaped routing in unidirectional narrow aisles.

Dividing every aisle into subsections, the time needed to travel through one wide aisle while applying a right-hand pick rule is given by the travel time per subsection $\tau_s$ multiplied by the distance $d$ of the subsection which is farthest away from the entrance of the aisle\footnote{We call $d$ the distance as it can be calculated by any norm which the user considers adapted. The easiest notion of distance is twice the maximum norm, so that $d = 2 \max  \lbrace x_j - x \rbrace $, where $x$ is the entrance point of the aisle and $x_j$ the $j$-th item to be picked in this aisle.}. In case of narrow aisles, the time needed to travel through one aisle is constant, namely just the number of subsections per aisle multiplied $\tau_s$. In other words, $d$ is a constant and not a variable.

The routing time $t_{r}$ for one bin is then calculated as
\begin{equation}\label{eq:time_routing}
  t_{r} = N_{\text{aisle}} \cdot \tau_{\text{aisle}} + d \cdot \tau_{s}
\end{equation}
where  $N_{\text{aisle}}$ is the number of aisles that have to be entered to collect all items in this batch, $\tau_{\text{aisle}}$ the time needed to change from one aisle to the next  and $\tau_s$ the time it takes to
move one subsection within one aisle.

One might ask why this heuristic should give a good result. In fact, considering an isolated routing optimisation, as described in the literature review above, a simple s-shaped heuristic does not always give the best results, even though it helps to avoid congestion. However, in our case, it is precisely the mix of a solution to the SLAP with implicit routing heuristics which will produce an optimal solution on the condition of this wide-aisle s-shaped routing heuristic.

\subsubsection{Multi-level picking}
The literature often distinguishes between low-level and high-level picking.
In low-level picking systems the picker can directly collect the items from the storage racks, while \defi{high-level picking} or ``\defi{man-aboard} order-picking'' indicates the use of a lifting order-pick truck or crane, see \cite{de2007design} for a detailed exposition. We design our algorithm in a way that the pick times depend on the level where the item is stored. In the simplest case, when only a distinction between low-level and high-level picking is made, we therefore work with two picking times, namely $ \tau_{l}$ for the lower level(s) and $\tau_{u}$ for ``upper level'' picking. The pick time per bin can then be calculated as
\begin{equation}
  \label{eq:timepick}
  t_{p} = N_{l} \cdot \tau_{l} + N_{u} \cdot \tau_{u} + \Theta(N_{u}) \cdot \tau_{\text{lift}}
\end{equation}
where $N_{l}$ is the number of items located in lower levels and $N_{u}$ the number of items located in the upper levels, respectively. % with the corresponding picking times $\tau_{u},\tau_{l}$.
The last term in equation (\ref{eq:timepick}) adds the time $\tau_{lift}$   needed to fetch or adjust the lifting device to the upper level(s). In the simplest situation of only one level change, $\Theta(N_{u})$ function returns one if there are any elements to retrieve from the upper levels and zero if the picker has no need to visit the upper levels.

Formula \eqref{eq:timepick} is only the special case of the general multi-level picking time formula
\begin{equation} =  \label{eq:multitimepick}
  \tilde{t}_{p} \; = \; (L-1) \cdot \tau_{\text{lift}} \sum_{j=1}^{L} N_{j} \cdot \tau_{j}
\end{equation}
where $L$ is the number of level changes to be made by the picker and $\tau_j$ the picking time for an item stored in level $j$. 

By specifying different picking times for different levels or other special situations, the proposed algorithm can be adapted to combined low- and high-level picking, and by adjusting $\tau_{\text{lift}}$, $\tau_{\text{aisle}}$  and $\tau_s$, the algorithm applies also to  differently shaped warehouses and variable routing heuristics.

\subsection{Construction of moves}

The total retrieval time constructed in \eqref{eq:totaltime} is the cost-function which our simulated annealing algorithm has to minimize. As explained above, such an algorithm  changes again and again the location of the storage containers (in which the items are stored), trying to find configurations with a lower total cost. These continuous changes are called \defi{moves}. It is of key importance for the performance of the algorithm to choose the correct type of moves.

The specific design our test warehouse adds an extra difficulty: the presence of different-size storage containers in the warehouse translate into combinatorial compatibility conditions, i.e. is has to be ensured  that the algorithm does not exchange two storage containers of different size. 

The crucial solution step here was to observe that the admissible combinations of container classes form subsections in the aisle. Therefore, in the presented algorithm, two update routines were implemented: the first routine exchanges two random boxes from the same size category, the second routine  swaps whole subsections in two randomly chosen levels. The first move is tasked to cluster all items that are strongly correlated and minimizes the individual item picking time for one bin. 
The second routine both  ensures that a solution found is admissible and accelerates the optimisation by searching for a more adequate location of  a group of already clustered items, e.g. a subsection consisting of items used to assemble a highly popular product option should be placed at a part of the warehouse which is quickly reachable by the picker. 

The combination of these two moves allows the simulated annealing algorithm to find warehouse configurations that are minimized by the average picking time for each bin, while keeping the overall performance picking time for the whole range of product options in mind.

\section{Validation and results}

\subsection{Case description}

We test our algorithm with real data from the production site of a medium-size European company offering highly customizable products with a long lifespan. When ordering a product, the customer chooses from a large number of possible options for the product of his choice, which are assembled at the production site. 

The individual parts for the product are prefabricated by subcontractors, shipped to the company and stored in the warehouse until needed, there is no just-in-time delivery. Orders to the warehouse arrive as a batch, which is picked in a single journey. The picker stores the items in the batch on a bin, which includes small trays for small pieces and a dedicated space for heavy items, so there is no issue of considering heavy or delicate items when deciding the routing strategy. When all items are collected, the picker delivers them to certain input points along the assembly line. 

The content of one batched order varies not only according to the output delivery point, which is the input point of a specific step in the assembly line, but also highly depends on the end-configuration of the product chosen by the customer. In other words, customised product options lead to complex correlations between the items stored in the warehouse. Positively correlated items are more likely to be found in one batched order arriving at the warehouse. 

\subsubsection{Sample warehouse design used in the algorithm}
Our algorithm was designed to be general enough to cope with several features appearing in the warehouse of the abovementioned company, which are variable aisle length, multiple storage levels, and different container types with combinatorial restrictions due to their size. 

\begin{figure*}
 \begin{center}
  \begin{tabular}{cc}
   \resizebox{0.5\linewidth}{!}{
    \includegraphics{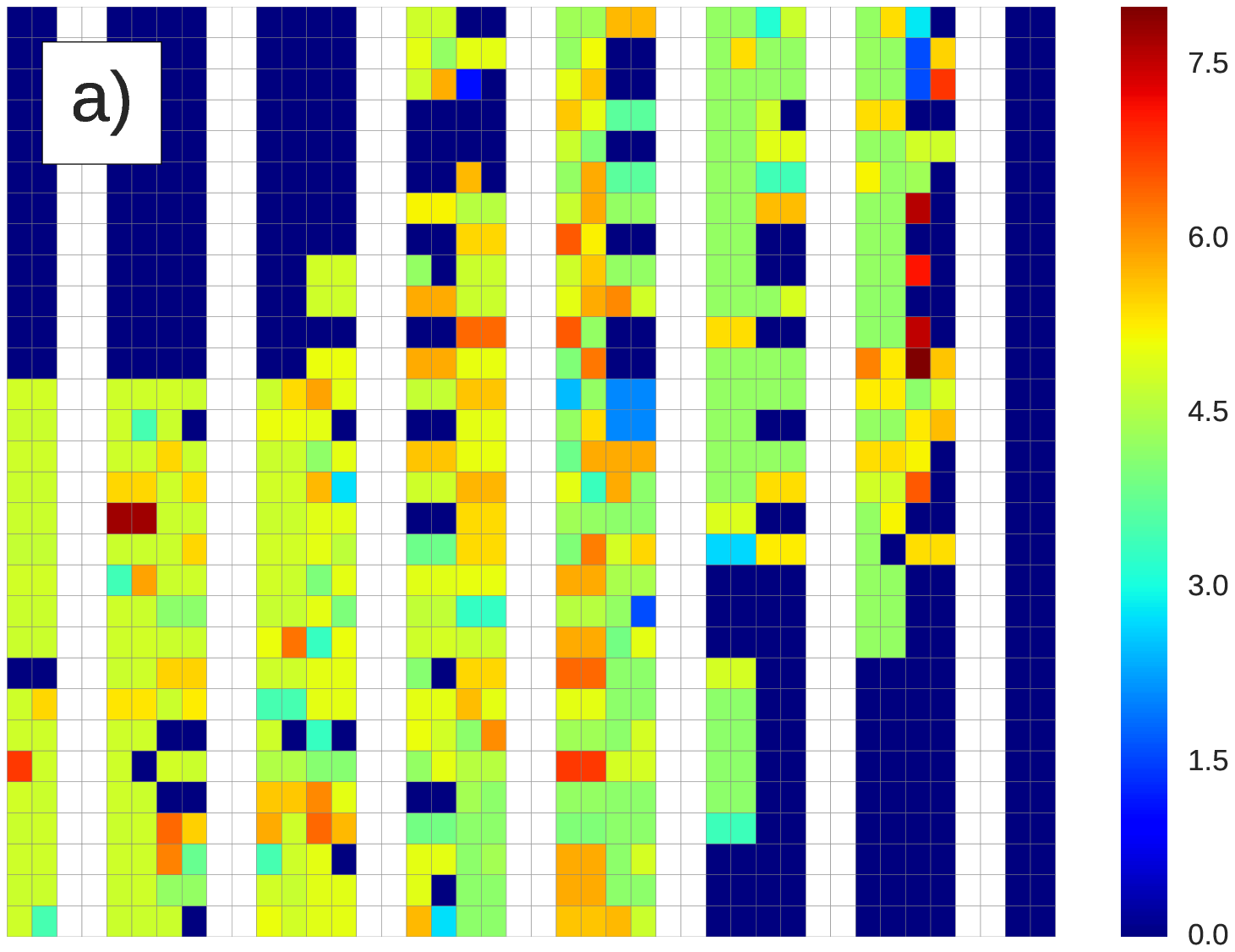}
   } & 
   \resizebox{0.5\linewidth}{!}{
    \includegraphics{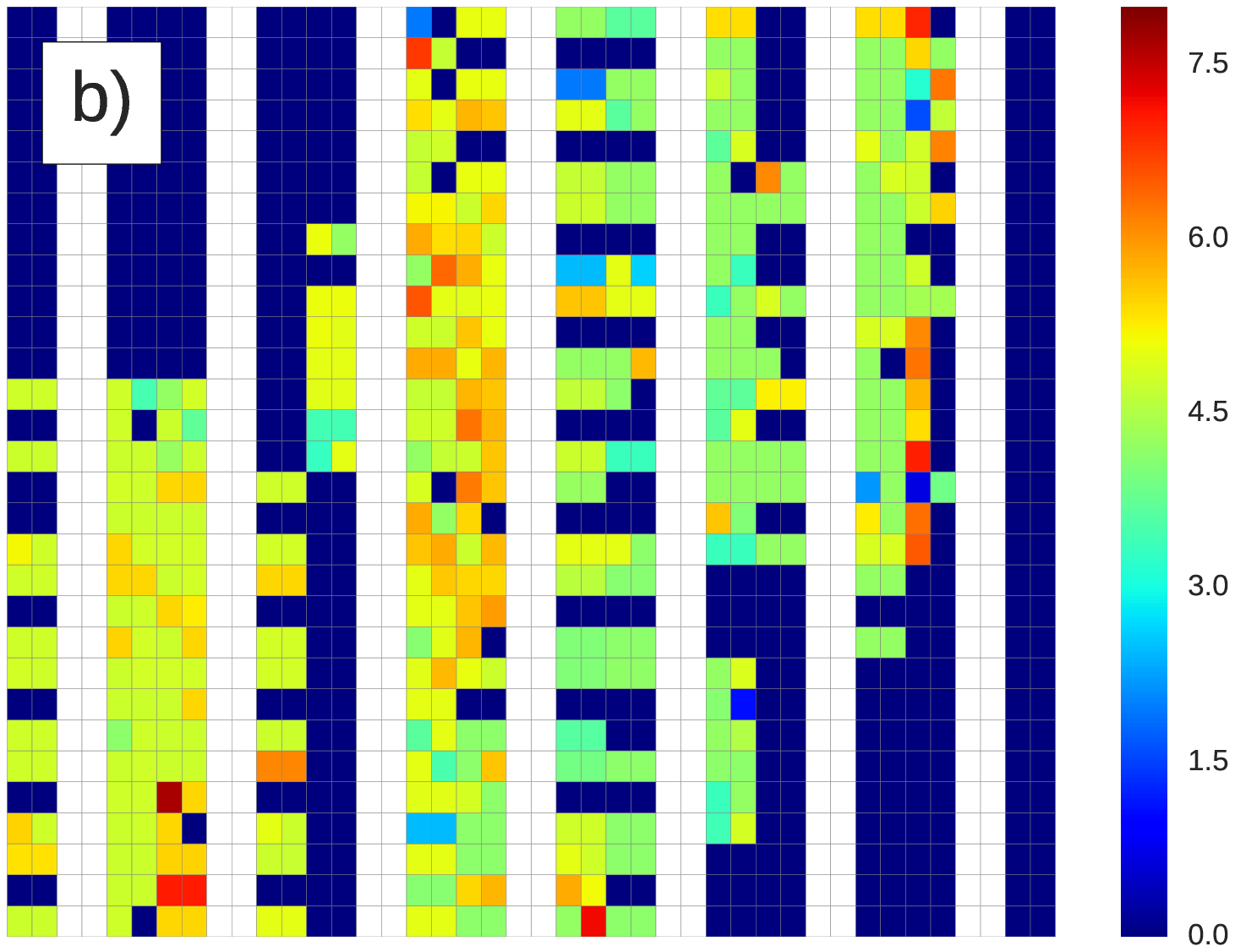}
   } \\
   \resizebox{0.5\linewidth}{!}{
    \includegraphics{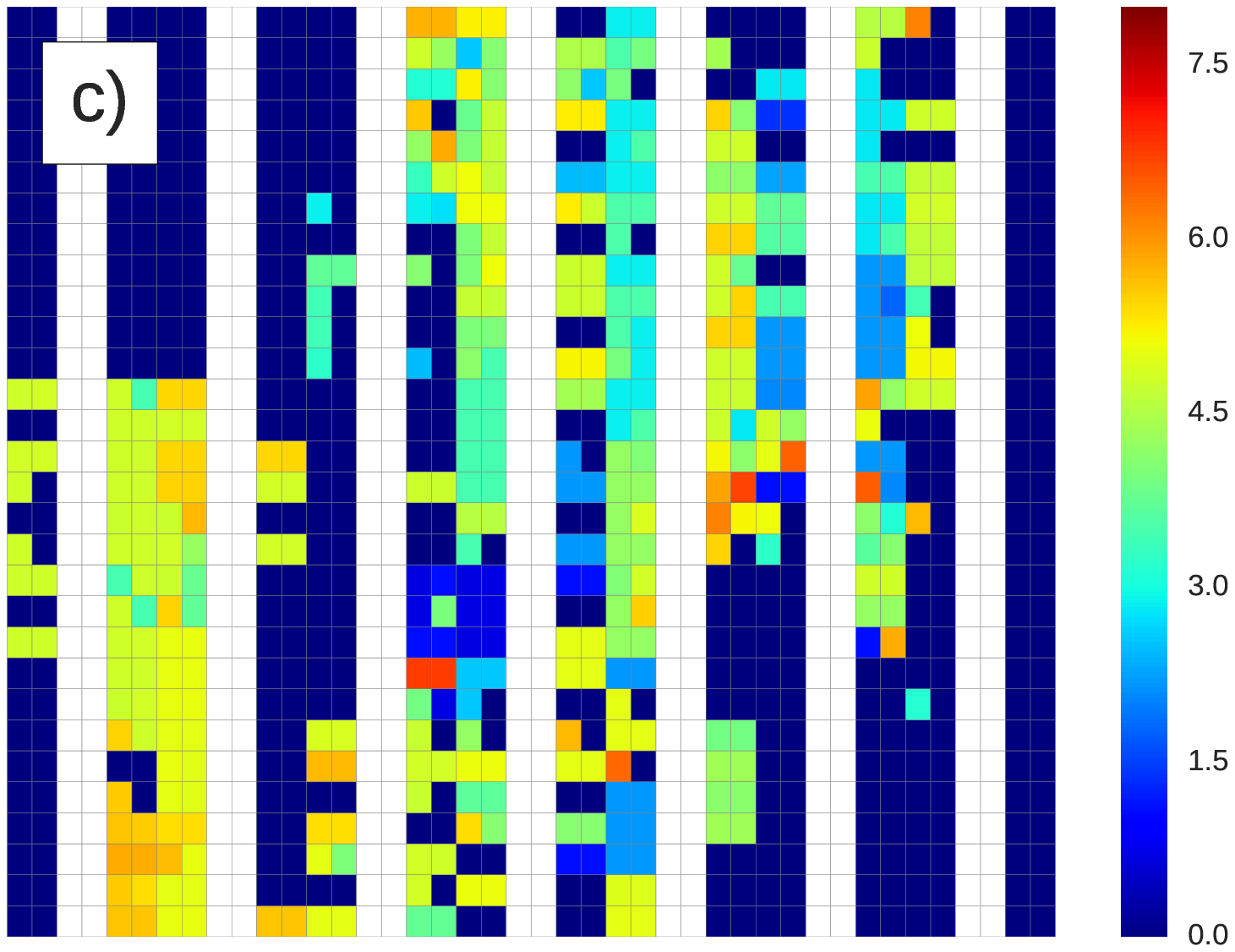}
   } & 
   \resizebox{0.5\linewidth}{!}{
    \includegraphics{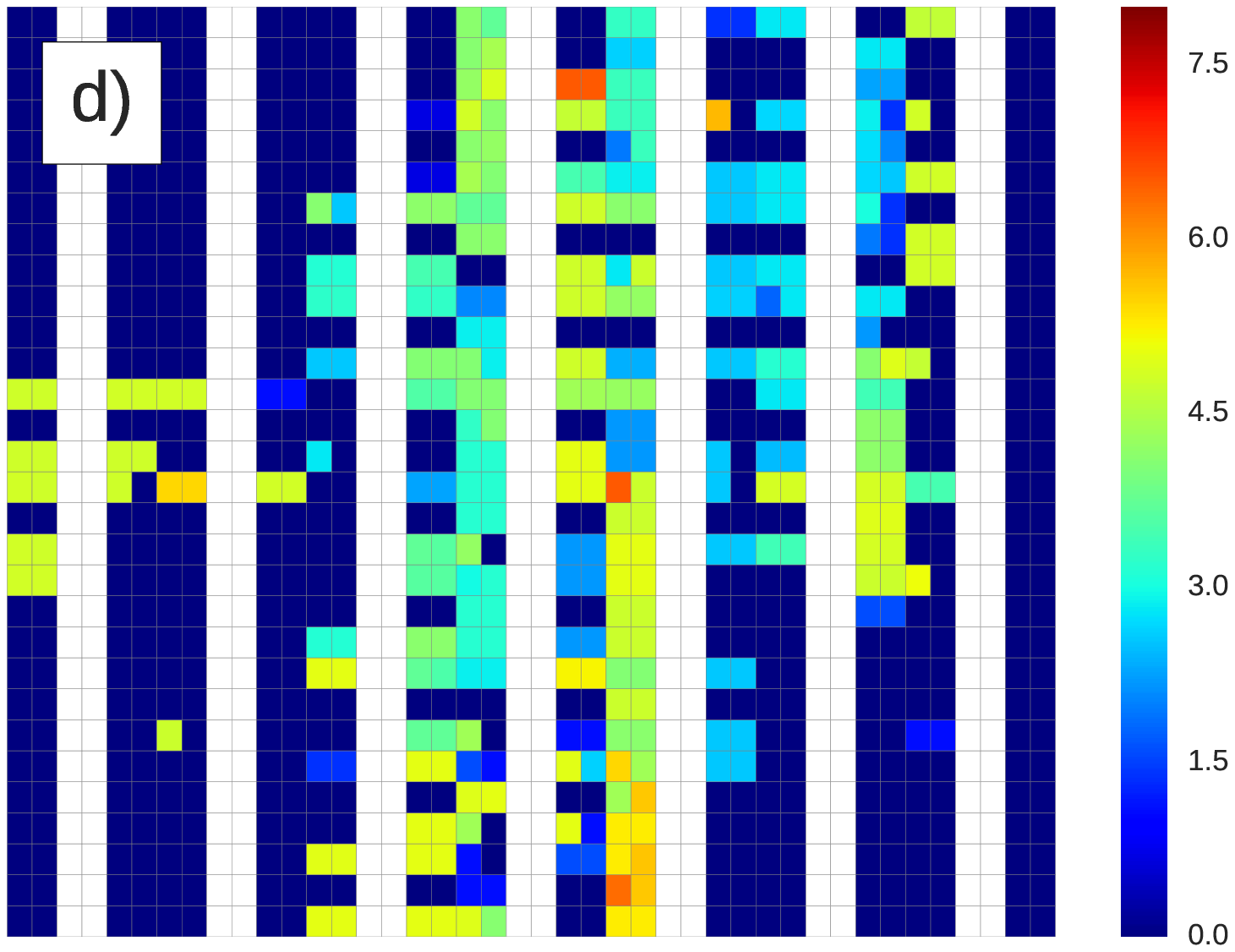}
   } \\
  \end{tabular}
 \end{center}
 \caption{Warehouse design divided into the different levels: a) ground level, b) first level, c) second level, d) third level.
Each pixel represents a designated storage space for one item. The colours indicate the logarithm of the total picking rate of an item.
 Red items are much more frequently picked than blue items. Pathways within the aisle are white. Dark blue indicates empty or non-exchangeable space due to real world constraints.}
 \label{fig:startLager}
\end{figure*}

As visualised in figure \ref{fig:startLager},  the sample warehouse of the manufacturer has 7 aisles of variable length (due to the constraints of the production site).
Each aisle is 4 levels high and is divided into (at most) 20 subsections.
Individual items are stored in containers, of which two main container type, namely large or regular size are used, and of these types, both high and low containers are available. Each subsection of an aisle is wide enough to hold two large or three regular containers, therefore up to six different items  (six items meaning three regular size containers of low height stacked on three other regular size containers of low height) can be stored at each level of a subsection. 

A total amount of 1268 different components are stored in the warehouse of the manufacturer whose data we used.
We considered only the primary pick location of these 1268 individual items, as only this primary pick location is mentioned in the order sheet given to the picker. The secondary location of this component is in the less frequented ``cold'' area of the warehouse, where a random storage policy is exibited (surplus and refill storage).

An order batch contains between one and 150 items each. As mentioned, the bin has designated storage options for different items. For example, small items go on the trays, so that no predefined sequence of picking is needed. The delivery locations are designated spots at the edge of the assembly line. As the retrieval time for one order batch is by orders of magnitude longer than the travelling time to the delivery points along the assembly line, the influence of those delivery points on the warehouse storage locations can be neglected.

\subsubsection{Routing heuristics in the company}

Our algorithm is designed to adapt to several routing situations. The routing heuristics used in our sample warehouse is as follows: pickers start with picking from the lower two levels of a wide aisle. They follow a right-hand rule, which means that they pick only on their right while traveling along the aisle. Once the last item to their right is reached, they turn around and pick the other side of the aisles until they  arrive back to the entrance of this aisle and change to the next aisle. After completing the routing in the lower levels,  a lifting order-pick truck is fetched and the picker starts picking the first upper level. 
After all items of the batch stored in the first upper level are picked, the height of the order-pick truck is adjusted and the picker continues the same routing on the second upper storage level. Using this routing heuristics, the company arranged their storage allocation based on the picking frequency. 

\subsubsection{Individual pick times}
Note that the time required for each pick (denoted by $\tau_j$ in formula \eqref{eq:multitimepick}) changes depending on the level from which items are picked. In our sample warehouse, we consider the simple situation of only two different picking times, $\tau_l$ for the lower levels, and $\tau_u$ for the upper levels, see formula \eqref{eq:timepick}. The time required for each pick in the upper levels is significantly longer than for the lower levels, as more careful steering is needed and the picker is less mobile.  
The choice of pick times and the time for aisle change/ level change used in the experiments of our algorithm are listed below. Please recall equations \eqref{eq:time_routing} and \eqref{eq:timepick} for the definition. 
It is worth mentioning that the solution quality does not depend on the exact times as long as they are in reasonable proportions
to each other.

\begin{center}
  \begin{tabular}{c|c}
  Variable & Time in seconds\\
  \hline
  $\tau_{aisle}$ & 30\\
  $\tau_s$ & 2 per subsection\\
  $\tau_l$ & 15\\
  $\tau_u$ & 30\\
  $\tau_{lift}$ & 120\\
  \hline
 \end{tabular}
 %\caption{Used times in equation \ref{eq:time_routing} and \ref{eq:timepick}}
 \label{tab:times}
\end{center}

\subsection{Validation of method}

 The experiments were carried out with 4192 pre-batched orders, representing the products assembled in the reference time frame.
All experiments were run single-threaded on Intel(R) Core(TM) i7-4790 CPU running at 3.60GHz and with 8 GB of memory.
We used a geometric cooling schedule starting at $T = 10^7$ and stopping when convergence is reached with a cooling constant $\alpha = 0.95$,
resulting in about $2 \cdot 10^6$ iterations. The total running time for one annealing was about 5 hours.

\begin{figure*}
 \begin{center}
  \begin{tabular}{cc}
   \resizebox{0.5\linewidth}{!}{
   \includegraphics{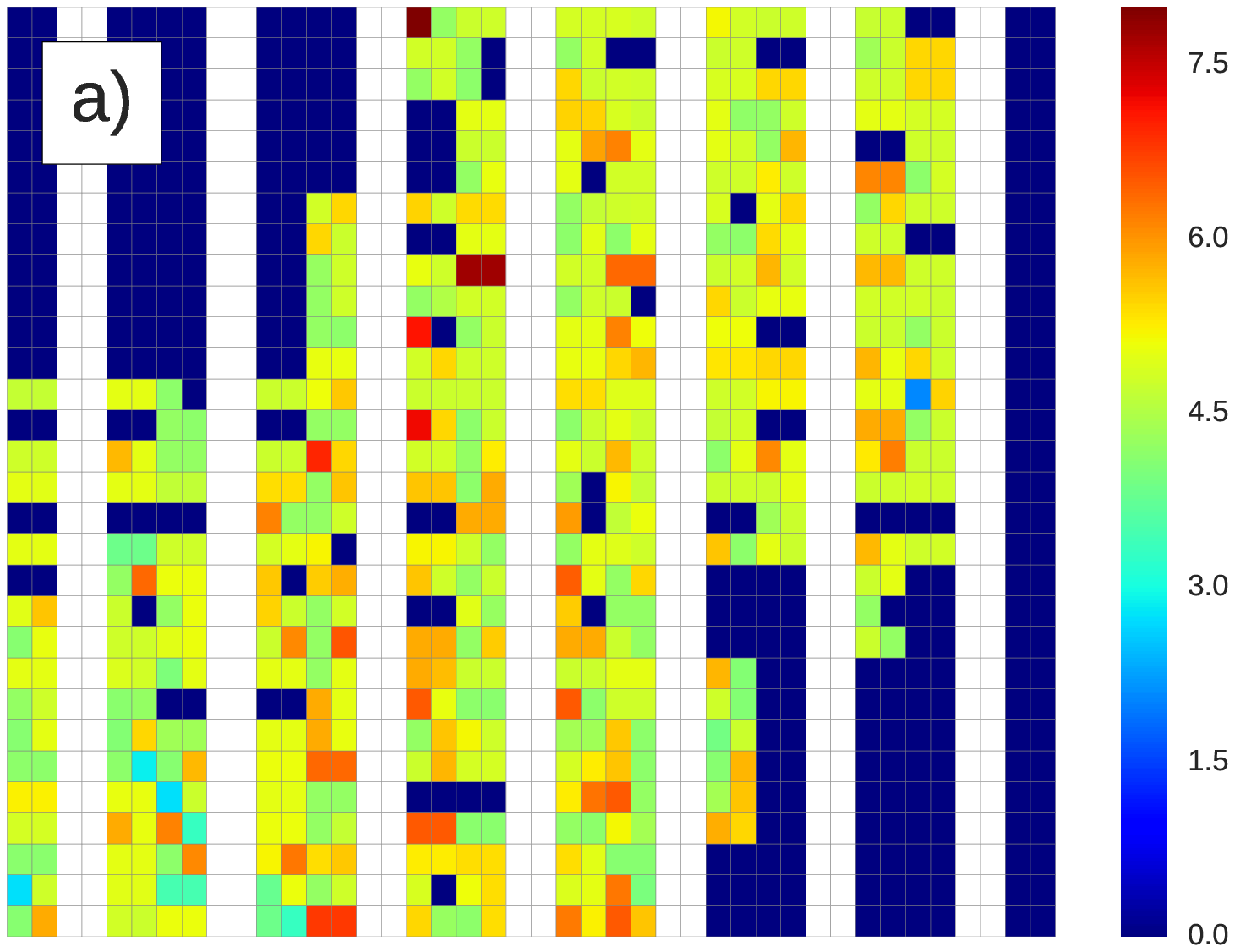}
   } & 
   \resizebox{0.5\linewidth}{!}{
    \includegraphics{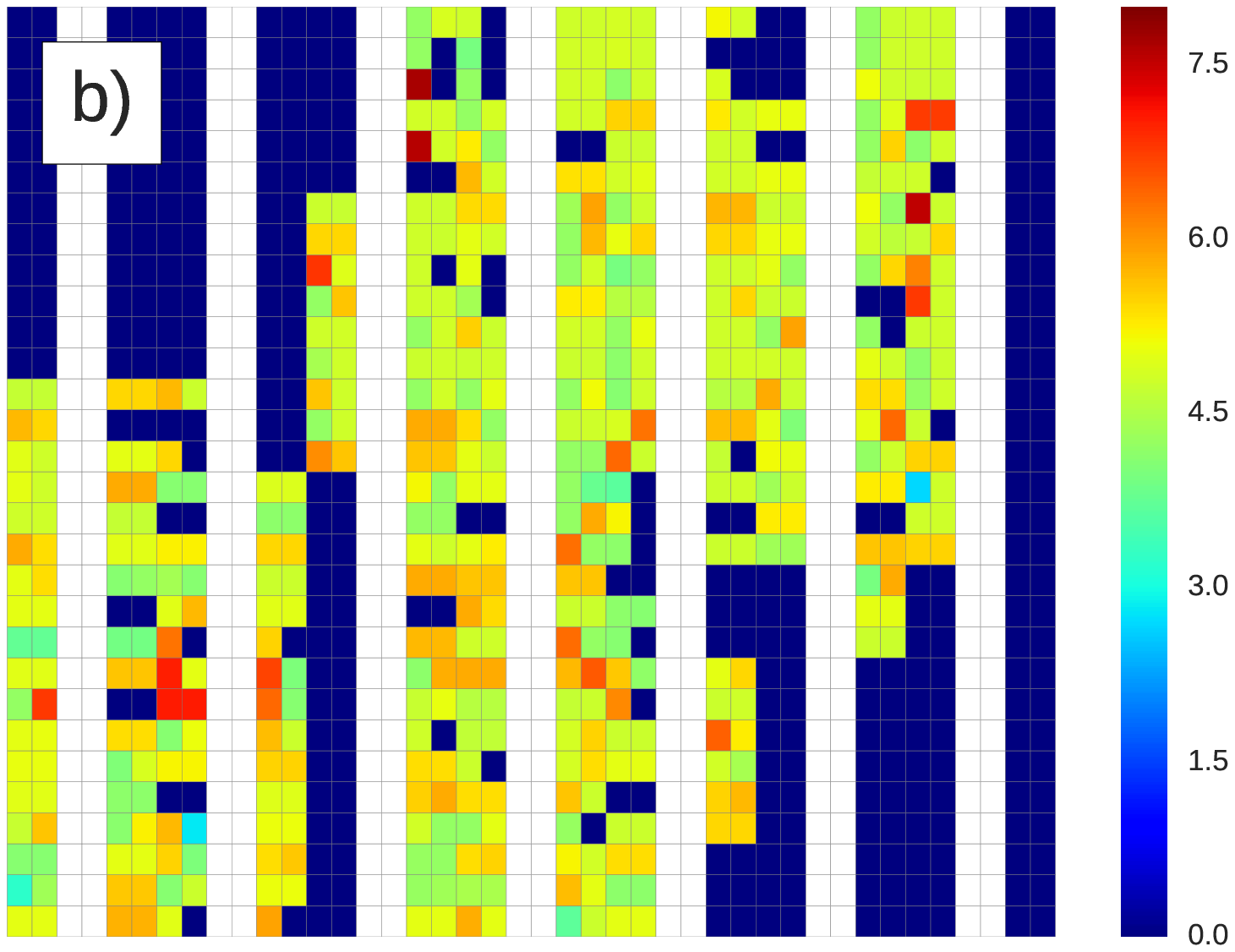}
   } \\
   \resizebox{0.5\linewidth}{!}{
    \includegraphics{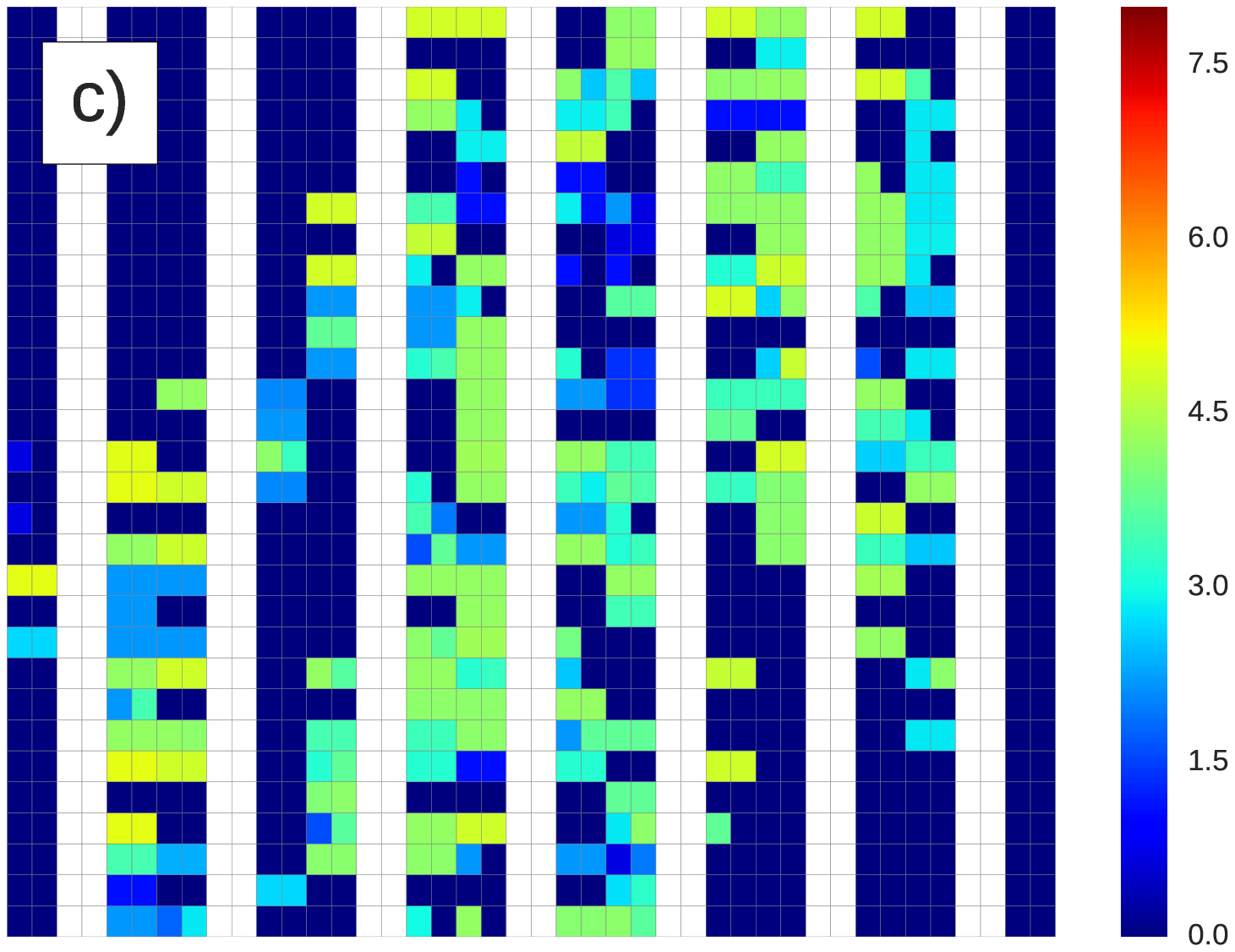}
   } & 
   \resizebox{0.5\linewidth}{!}{
    \includegraphics{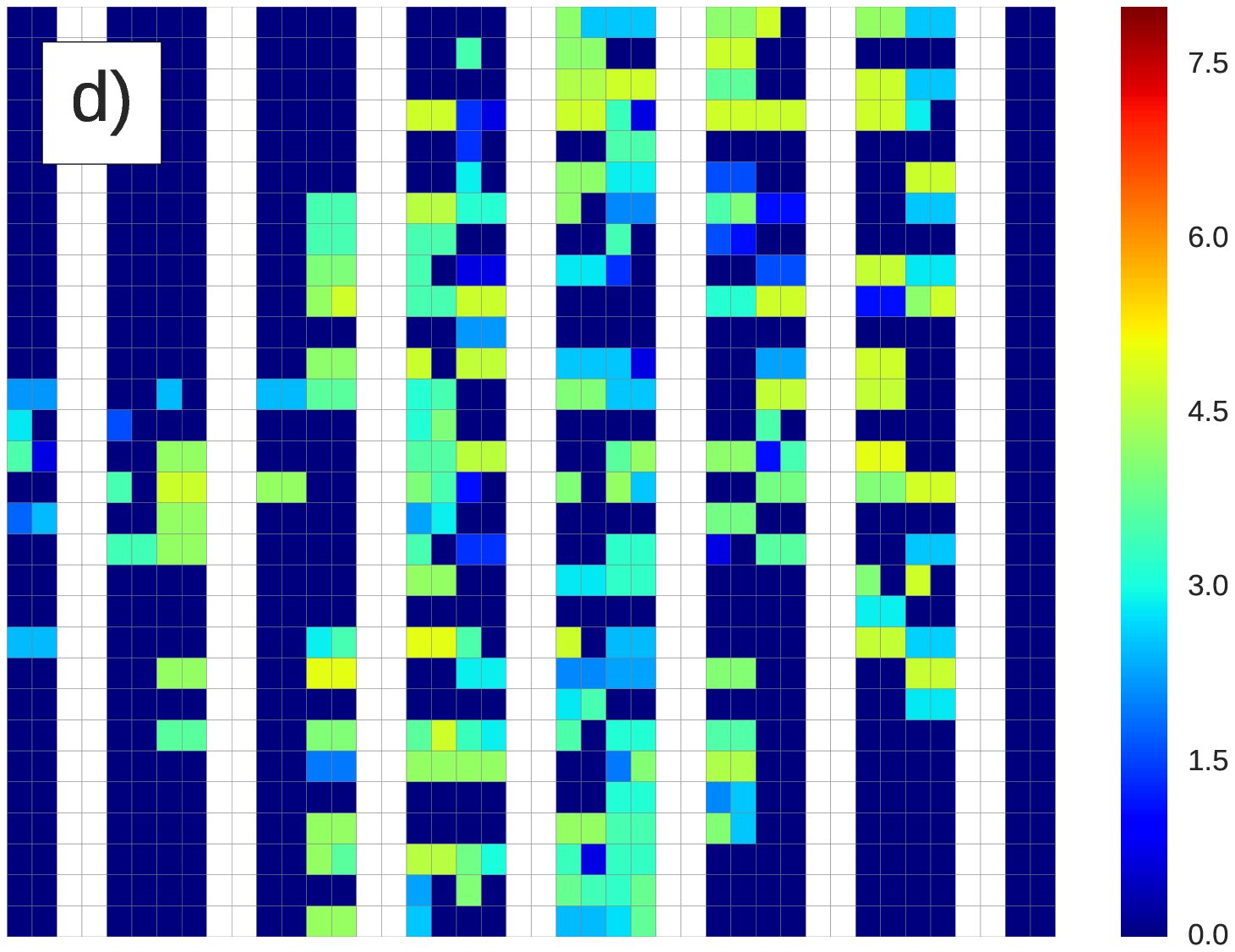}
   } \\
  \end{tabular}
 \end{center}
 \caption{Optimised warehouse design divided into the different levels: a) ground level, b) first level, c) second level, d) third level.
 The colours indicate the logarithm of the total picking rate of an item.
 Red items are much more frequently picked than blue items. Pathways within the aisle are white. Dark blue indicates empty or non-exchangeable space due to real world constraints.}
 \label{fig:optLager}
\end{figure*}
%Assuming a random item distribution in the warehouse, this simulated annealing algorithm reduces the overall picking time approximately by 38\%. 
\subsection{Results}
After  optimisation of storage locations  by our algorithm, the total retrieval time for all $4 \cdot 10^4$ pre-batched orders in the reference time frame was calculated to be  approximately 38\% lower than for a random item distribution and  21\% lower compared to the total retrieval time with the pre-otimised storage location configuration used by the company. 

This is a significant improvement, in particular as the initial storage allocation used a frequency-based heuristics similar to \cite{battista2011storage}: Fixing the routing heuristics, the company's logistics division had already re-arranged the items used more often to the first aisles and those used less were put in the back aisles.  However, as our results show, this simple heuristics has its drawback when handling large batched orders.

\subsubsection{Detailed analysis - global distribution}
Figure \ref{fig:optLager} shows the optimised warehouse design. Red items have the highest picking frequency, blue items the lowest. %\textcolor{green}{The scale is given by the logarithm of the frequency plus one.} 
In comparison to the starting situation (Figure \ref{fig:startLager}), the lower levels shown in a) and b) are better stocked. They contain the more frequently picked items (red, orange, yellow/green pixels), while the upper levels c) and d)) contain
more blue/green items. In fact, the algorithm removed all orange/red items from the upper level, which is to be expected since the higher picking time in the top floors and the need for a lifting device have a significant impact on the optimisation.

The canonical entrance to the warehouse is on the right-bottom corner of the picture. The first aisle from the right, which is the first aisle visited, initially had a lot of very frequently picked items on the first two levels, as this was the preferred storage location in the heuristics used by the company. After optimisation by our algorithm, most of these items were moved, as the aisle is very short one and therefore inefficient to visit from a global optimisation point of view, as taken by our algorithm.  

\subsubsection{Detailed analysis - batch-induced correlations} 
As already discussed above, the picking process of large batched orders results in complex correlations between the individual items in the warehouse. The correlations are not necessarily related to the frequency of picks of a single item: some items are very basic and are used for every single product that this company is producing,  independently of the product option chosen by the customer, while other items are specific to one product option and appear in exactly one batch if and only if this product option was ordered.

One might conjecture that clustering the items according to their correlation to each other leads to a lower total retrieval time.
By clustering we mean that highly correlated items are stored in neighbouring storage containers.  

\begin{figure*}
	\begin{center}
		\begin{tabular}{cc}
			\resizebox{0.5\linewidth}{!}{
				\includegraphics{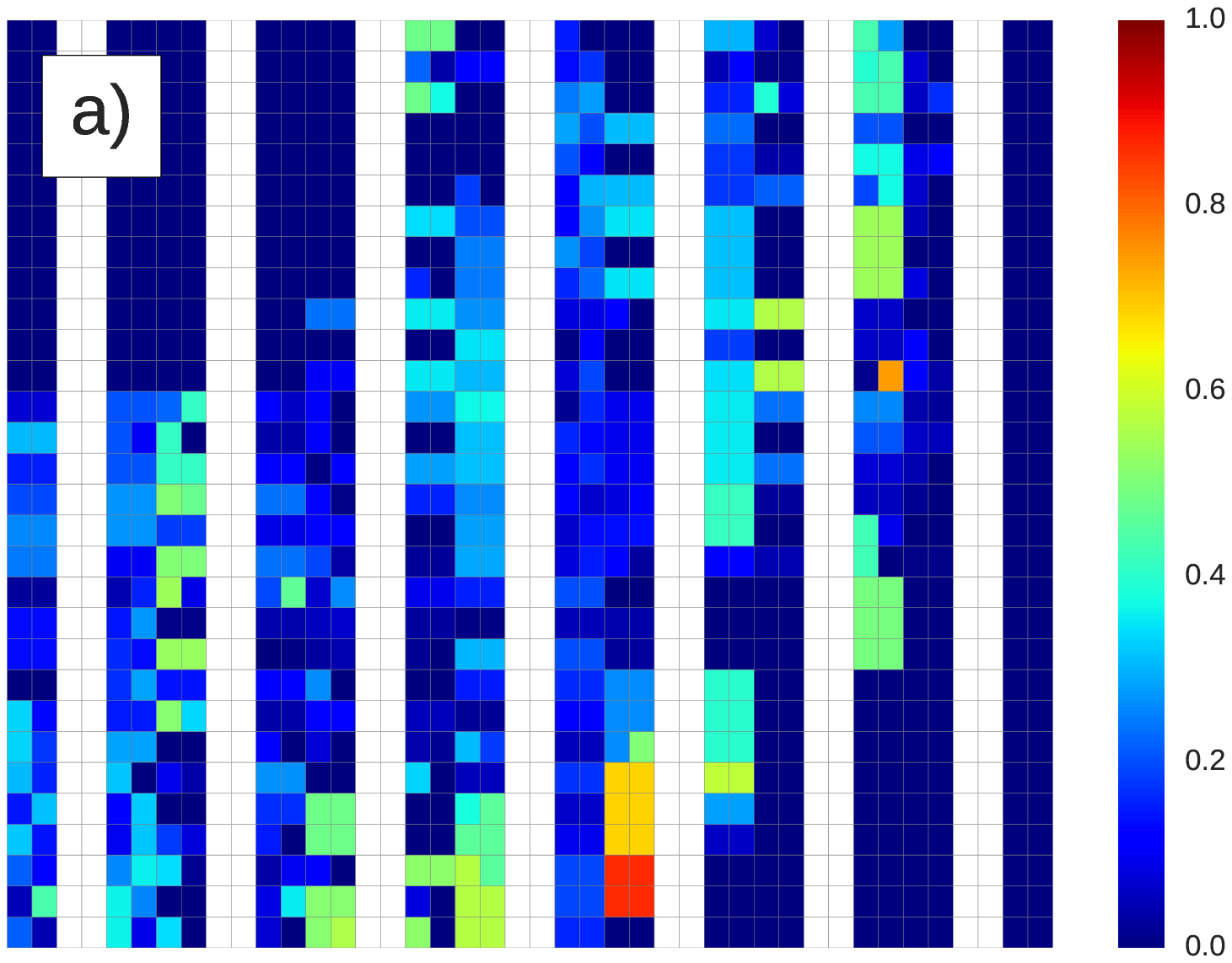}
			} & 
			\resizebox{0.5\linewidth}{!}{
				\includegraphics{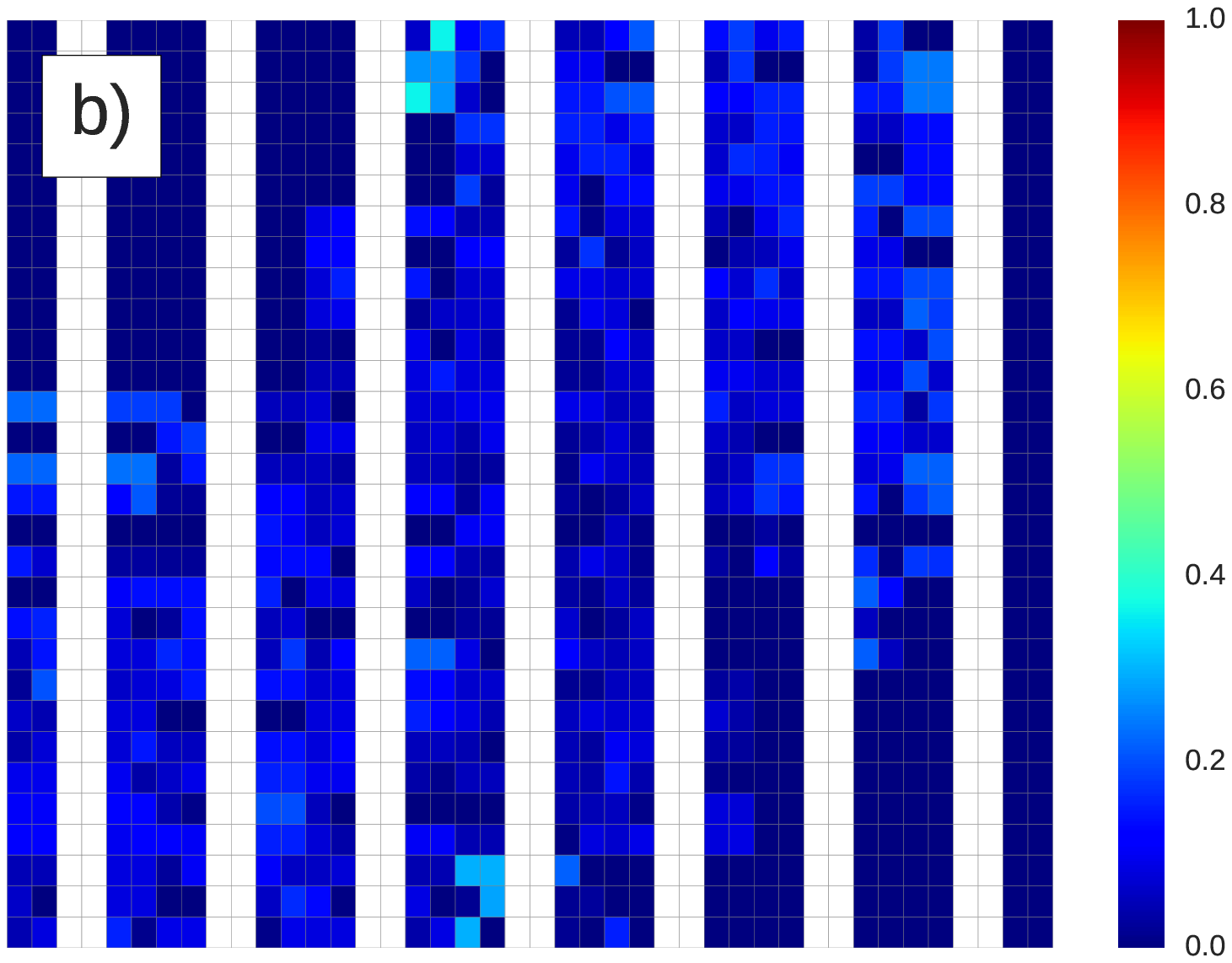}
			}
	\end{tabular}
	\end{center}
	\caption{The average jaccard-similarity for each item in the lowest level a) before b) after optimisation}
	\label{fig:corr_lager_e0}
\end{figure*}

To check if our algorithm does indeed cluster correlated items, we visualise in figure \ref{fig:corr_lager_e0}  the changes in correlation between neighbouring items. The visualisation is done via the average jaccard-similarity coefficient of a batch which contains item $i$ and a batch which contains a direct neighbour of item $i$,which we call $j$.  Roughly speaking, a high jaccard-similarity coefficient  means that item $i$ and its neighbour $j$ are positively correlated in the picking process, i.e. a large percentage of batches which contain $i$ also contain $j$.

The calculation of the jaccard-similarity coefficient goes as follows:
Denote $\left\{B_i\right\}$ a batch in which $i$ occurs and $\left\{B_j\right\}$ a batch in which $j$ occurs.  The jaccard-similarity coefficient measures the ``similarity'' between two finite sample sets $\left\{B_i\right\}$ and $\left\{B_j\right\}$  and is defined by
\begin{equation}
\mathrm{sim}\left(\left\{B_i\right\},\left\{B_j\right\}\right)=\frac{\left|\left\{B_i\right\}\bigcap\left\{B_j\right\}\right|}{\left|\left\{B_i\right\}\bigcup\left\{B_j\right\}\right|}.
\end{equation}
Moreover, define the set of neighbours of item $i$ as the set of  items $j$, which are in the same or adjacent subsection of an aisle. We require that $j$ has to be stored in the same level category as $i$, meaning that if $i$ is stored in a lower level, then also $j$ has to be stored in a lower level to qualify as a neighbour. 
%Then, for each item $i$ in the warehouse, the correlation between neighboring items can be 
%we calculate the average jaccard-similarity between the set of batches $\left\{B_i\right\}$ and the set of batches $\left\{B_j\right\}$ is calculated. 

Figure \ref{fig:corr_lager_e0} shows  that the heuristic approach taken by the company (plot  a) of figure \ref{fig:corr_lager_e0}) shows a high similarity between a few items, i.e. objects which are often picked to the same batch are stored in neighbouring containers. However,  the optimised configuration(plot b) of figure  \ref{fig:corr_lager_e0}) does not display areas of a high similarity.

Consequently,  in contrast to intuition, clustering of correlated items does not necessarily lead to a reduction in retrieval times. This phenomenon can be explained with the relatively low impact of item distance to the total retrieval time of a large batch, in relation to the sum of the picking times of the individual items.
This behaviour is also present in the other warehouse levels.
 
To conclude, the analysis of correlations of neighbouring items shows that the examined warehouse cannot be simply divided into  ``hot areas'' and a ``cold areas'' with respect to batched orders:  the complex structure of the batches leads to highly non-trivial optimised configurations.

\begin{figure*}
\begin{center}
	\resizebox{\linewidth}{!}{
		\includegraphics{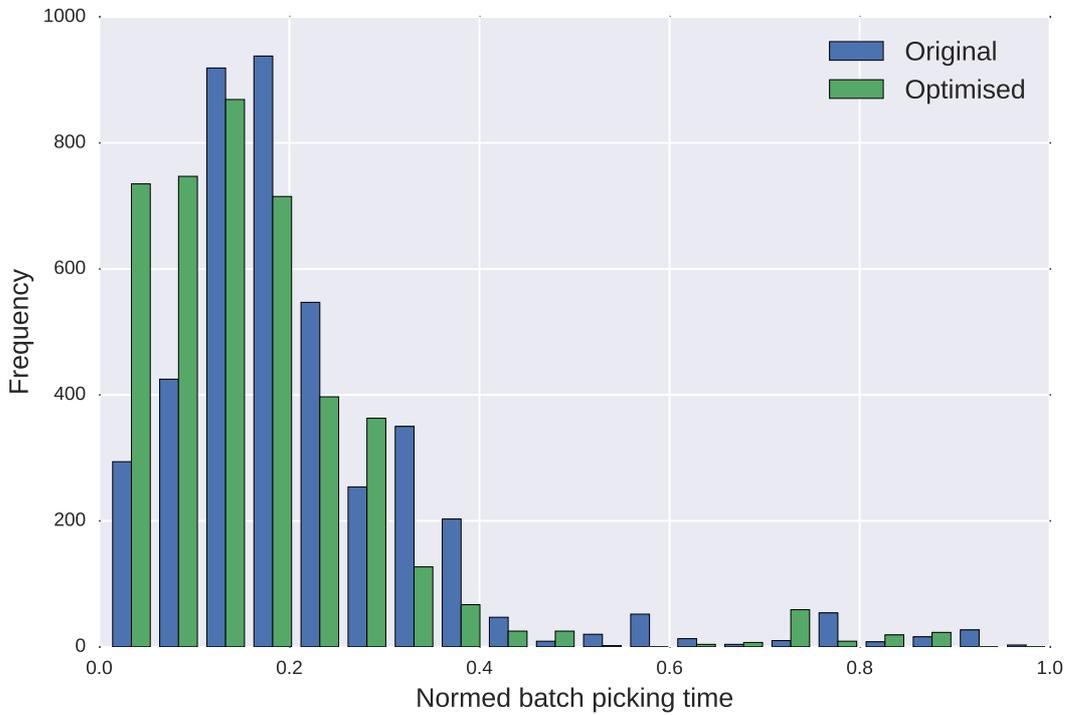}
	}

\caption{Histogram of individual batch picking times normed to the maximum of the original picking time (blue) before and  (green) after optimisation.}
\label{fig:hist_picking}
\end{center}
\end{figure*}

\subsubsection{Detailed analysis - individual picking times}
Another important question is whether an optimal total retrieval time for a large number of batches might go to the expense of very long retrieval times for a few batches.
The last step in our analysis is therefore to check the individual picking times per item. For better visibility, the displayed picking times are normed to the maximum batch picking time in the original configuration.
The histogram of individual picking times are shown in figure \ref{fig:hist_picking}: Before the optimisation, the distribution of individual picking times is centred  around a value of $0.16$ with a right handed fat tail.
The algorithm is able to shift the distribution to shorter times and eliminates parts of the fat tail. This proves that the optimal storage assignment obtained by our algorithm bears a lower probability  of extremely long retrieval times for arbitrary batches.

\section{Discussion/Conclusion}

This paper presents a simulated annealing algorithm to reduce picking times of potentially large sets of batched orders by a combined optimisation of the storage assignments in a multi-level warehouse and an implicit s-shape routing heuristic.

Our main contribution is a general solution method for the solution of the storage location assignment problem under adaptable routing methods, which is flexible enough to be used in a multi-level warehouse setting. While the current algorithm was designed for single-block warehouse layouts, extensions to more general warehouse settings can easily be done by adapting formula \eqref{eq:time_routing}. Notably, this algorithm can also be applied to parts of the warehouse without resulting in phantom constraints or other unnatural solutions to this combined optimisation problem.

To our knowledge, our article is the  first to investigate the impact of large batched orders on the optimal storage assignment. Contrary to intuition, we give evidence that, even for batched orders of very large size (e.g. over 100 items in one batch), clustering of correlated items in specific parts of the warehouse does not necessarily lead to a reduction in retrieval times. Moreover, we show that clustering heuristics are not even beneficial to reduce the probability of extremely long retrieval times for rarely occurring batches.  % (see Supplementary).
%complex entanglements between the items to be picked

We  tested our algorithm on real data, optimising the storage assignments in a four-level warehouse of a manufacturer with pre-batched orders of 1-150 items per order. The optimal storage assignment suggested by the algorithm reduces the total retrieval time by 21\% compared to heuristics based on the frequency of picking for individual pieces.
%The total picking time was reduced by 21\% in comparison to the original warehouse configuration.
Assuming a random item distribution in the warehouse, this simulated annealing algorithm reduces the overall picking time approximately by 38\%. These savings are achieved without changing the routing heuristics and without splitting large pre-batched orders. The simultaneous delivery of all items, even of a very large batch, is crucial to maintaining efficiency of the assembly line. Note that the algorithm is very fast, as all parts of the algorithm are designed and optimised for multi-level warehouses. No black-box packages are used.

The main novelty of our method is to provide a structured approach for the storage location assignment problem with very general pre-batching constraints  in multi-level warehouses settings. In particular, the algorithm is able to deal with complex correlated batches and able to find highly non-trivial optimised configurations.

\section*{Acknowledgements}

We gratefully acknowledge the financial, data and feedback contribution from our partner company.  A.E. and C.G. thank Stiftung der Deutschen Wirtschaft for partial funding of this project. C.G.'s research is supported by ERC grant no. 277749 ``EPSILON''.

\bibliographystyle{abbrv}   % this means that the order of references
			    % is dtermined by the order in which the

			    % \cite and \nocite commands appear
%\bibliography{hako}  % list here all the bibliographies that
			     % you need.

\end{document}